%% file: main.tex
\PassOptionsToPackage{table}{xcolor}
\documentclass[runningheads]{llncs}

\usepackage{eccv}

\input{preamble}

\usepackage{eccvabbrv}

\usepackage[accsupp]{axessibility}  

\usepackage{xr-hyper}

\usepackage{hyperref}

\usepackage{orcidlink}

\begin{document}

\title{
    \ourmodel: Metric Feed-Forward 3D Reconstruction Prior for Multi-view 3D Object Detection from Streaming Inputs
}

\titlerunning{\ourmodel}


\author{
    Yung-Hsu Yang\inst{1}\orcidlink{000-0003-0044-515X} \and Luigi Piccinelli\inst{1}\orcidlink{0009-0002-7848-7325} \and Samuel Rota Bulò\inst{2}\orcidlink{0000-0002-2372-1367} \and Sunghwan Hong\inst{1}\orcidlink{0000-0003-0685-3779} \and \\
    Denis Rozumny\inst{2}\orcidlink{0000-0001-9874-1349} \and Johannes Schönberger\inst{2} \and Zuria Bauer\inst{1}\orcidlink{0000-0001-8447-2344} \and Hermann Blum\inst{3}\orcidlink{0000-0002-1713-7877} \and \\
    Peter Kontschieder\inst{2} \and Marc Pollefeys\inst{1}\orcidlink{0000-0003-2448-2318}
}

\authorrunning{Y-H.~Yang et al.}


\institute{ETH Zürich, Switzerland \and Meta Reality Labs Zürich, Switzerland \and University of Bonn, Germany}

\maketitle

\input{sec/00_Abstract}
\input{sec/01_Introduction}
\input{sec/02_Related}
\input{sec/03_Method}
\input{sec/04_Experiments}
\input{sec/05_Conclusion}

%
%
\bibliographystyle{splncs04}
\bibliography{main}



\end{document}

%% file: preamble.tex
\usepackage{acro}
\usepackage{mathtools}
\usepackage{graphicx}
\usepackage{booktabs}
\usepackage{multirow}
\usepackage{wrapfig}
\usepackage{lipsum}

\newcommand*{\ourmodel}{Map-Det3D\@\xspace}

\definecolor{colorFst}{HTML}{bde6cd}
\definecolor{colorSnd}{HTML}{e4eebc}
\definecolor{colorTrd}{HTML}{fff8c5}

\newcommand{\parsection}[1]{\vspace{5pt}\noindent\textbf{#1}\xspace}
\newcommand{\PAR}[1]{\vspace{3pt}\noindent\textbf{#1}\xspace}

\newcommand{\video}{\mathcal{V}}
\newcommand{\img}{\mathbf{I}}
\newcommand{\camK}{\mathbf{K}}
\newcommand{\camP}{\mathbf{P}}
\newcommand{\cam}{\mathcal{C}}
\newcommand{\nviews}{T}
\newcommand{\winsize}{T}
\newcommand{\window}{\mathcal{W}}

\newcommand{\detset}{\mathbf{B}^{\text{3D}}}
\newcommand{\detscores}{\boldsymbol{\sigma}}

\newcommand{\detboxtwoD}{\mathbf{B}^{\text{2D}}}

\newcommand{\scale}{\rho}

\newcommand{\featenc}{\mathbf{F}_{\text{E}}}
\newcommand{\msfeats}{\mathcal{F}}
\newcommand{\npatches}{P}
\newcommand{\qimg}{\mathbf{Q}^{\text{IMG}}}
\newcommand{\qbox}{\mathbf{Q}}
\newcommand{\qscale}{q_{\text{scale}}}

\newcommand{\nlayers}{L}
\newcommand{\nqueries}{M}
\newcommand{\ffr}{\Phi_{\text{FF3R}}}
\newcommand{\detmod}{\Phi_{\text{DET}}}
\newcommand{\headthreeD}{\Phi_{\text{3D}}}

\newcommand{\Ltotal}{\mathcal{L}_{\text{total}}}
\newcommand{\LtwoD}{\mathcal{L}_{\text{2D}}}
\newcommand{\LthreeD}{\mathcal{L}_{\text{3D}}}
\newcommand{\Lxy}{\mathcal{L}_{xy}}
\newcommand{\Lz}{\mathcal{L}_{z}}
\newcommand{\Ldim}{\mathcal{L}_{\text{dim}}}
\newcommand{\Lrot}{\mathcal{L}_{\text{rot}}}

\usepackage{microtype}

%% file: sec/00_Abstract.tex
\begin{abstract}
Metric 3D object detection is a core capability for embodied agents, yet most reliable systems lean on depth sensors, trading away cost, power, and integration simplicity.
This motivates monocular 3D detection, which avoids additional constraints, yet it faces a major obstacle: from a single image, depth, and especially absolute scale, are underconstrained.
As a result, the prevailing pattern of detecting in 2D and then predicting 3D attributes is often brittle, since modest range errors can dominate 3D localization, and the learned scale prior can fail when cameras, motion, or environments undergo domain shifts.
To address this, we propose \ourmodel, an online multi-view 3D object detection model that brings detection directly into a 3D space reconstructed from RGB.
We map a short temporal window into multiple views and repurpose a feed-forward metric 3D reconstruction model as our geometric backbone while tuning its object-aware capabilities.
Building on this representation, \ourmodel directly predicts boxes in metric 3D space, without the widely used 2D-to-3D lifting.
Experiments across different benchmarks show that this design supports strong online performance and robust transfer without adaptation, suggesting that training reconstruction priors for detection is a practical route to stable metric 3D detection from monocular video.
Code and models are available at \href{https://royyang0714.github.io/Map-Det3D}{royyang0714.github.io/Map-Det3D}.

\keywords{Multi-view 3D Object Detection \and Scene Understanding  \and Feed-Forward Metric 3D Reconstruction}
\end{abstract}

%% file: sec/01_Introduction.tex
\begin{figure}[t]
    \centering
    \footnotesize
    \includegraphics[width=1.0\linewidth]{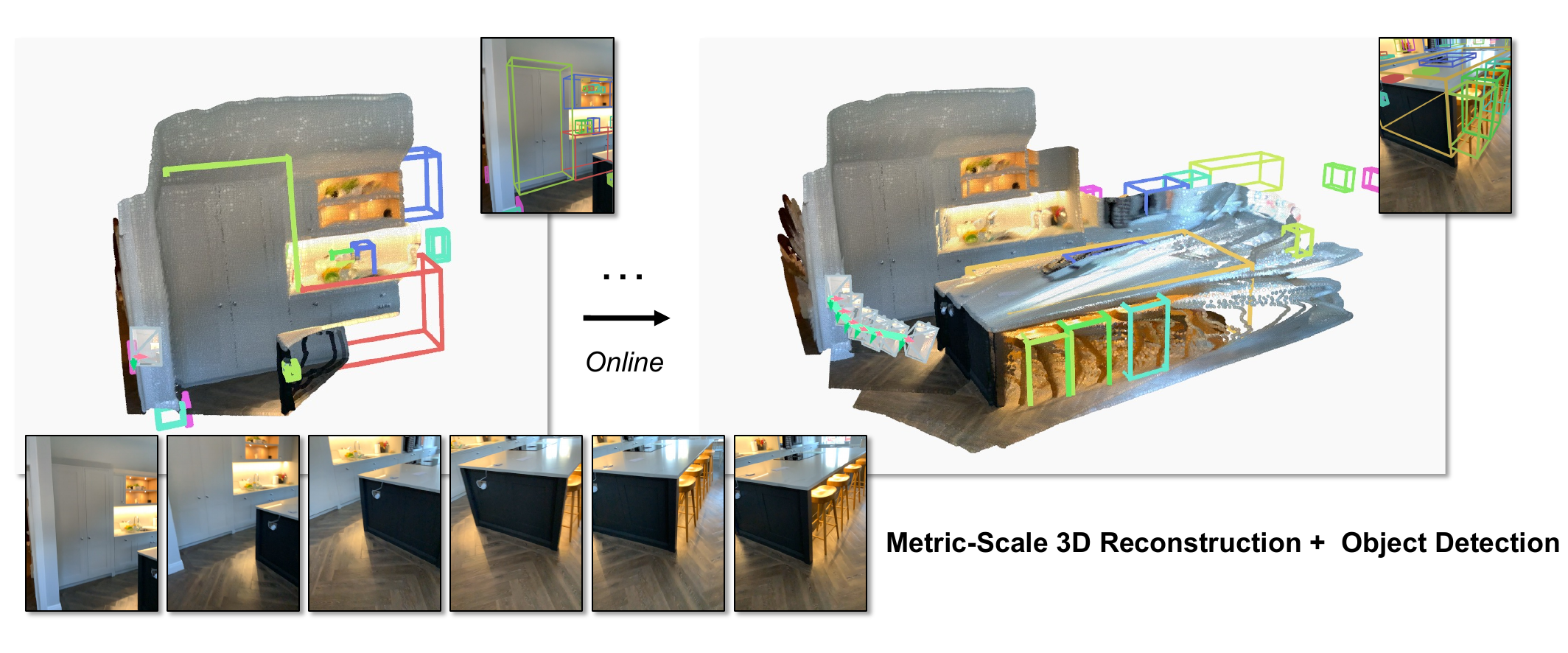}
    \caption{
        \textbf{\ourmodel.}
        We leverage the metric-scale geometry prior from a multi-view feed-forward 3D reconstruction model and adapts it for 3D object detection from images. 
        In particular, given a streaming video input, \ourmodel can perform online metric-scale 3D reconstruction and 3D object detection simultaneously in the form of feed-forward inference over a sliding window. \ourmodel can run on videos alone, but camera poses and intrinsics can be added to further stabilize the geometry estimation.
    }
    \label{fig:teaser}
\end{figure}

\section{Introduction}
\label{sec:intro}
3D object detection (3DOD) is a fundamental capability for embodied AI, autonomous navigation, and augmented reality, where agents must reason about objects in metric 3D to act safely and effectively.
In many deployed systems, this problem is made substantially easier by active depth sensors such as LiDAR or RGB-D cameras, which directly provide 3D geometry.
However, these sensors come with practical drawbacks: prohibitive hardware costs, high power consumption, and restrictive form-factor constraints that are difficult to satisfy on lightweight and consumer platforms.
As a result, enabling accurate 3D object detection from images alone has become an increasingly desirable alternative.

A long-standing challenge is that, from a single image, depth and its absolute scale are inherently ambiguous.
Consequently, monocular 3DOD methods~\cite{brazil2023omni3d,yang20253dmood,zhang2025detect,yao2024open,yao2026labelany3d} typically follow a 2D-to-3D lifting recipe: they detect objects in the image plane and then regress 3D attributes such as depth, size, and orientation using learned priors.
While effective in-distribution, this strategy is fragile in practice because small depth errors can dominate 3D overlap, and the learned scale prior may break under shifts in camera intrinsics, ego-motion patterns, or scene statistics.
These issues become particularly pronounced when transferring across devices and environments, precisely the setting where image-only 3DOD is most attractive.

An appealing alternative is to perform detection directly in a 3D representation, where metric reasoning about size, occlusion, and spatial extent is explicit.
In principle, operating in 3D mitigates scale ambiguity and reduces the burden on priors tied to the image plane.
In real-world scenarios, however, 3D geometry is often noisy, incomplete, and sensor-dependent: point clouds vary widely in density, scale, and noise characteristics, and models trained on one sensing regime can fail to generalize to another.
Moreover, compared to image-based pipelines, access to appearance cues is more indirect in 3D-only processing, and designing efficient mechanisms that provide 3D models with rich appearance information remains an active topic~\cite{zhang2025concerto,abouzeid2025ditr}.

Multi-view geometry offers a natural way to resolve single-view ambiguities by aggregating evidence across viewpoints.
Recently, feed-forward 3D reconstruction (FF3R) has emerged as a powerful source of geometric inductive bias~\cite{wang2025vggt,wang2025pi3,keetha2026mapanything,lin2025da3}.
Given a handful of input views, these models can robustly recover camera motion and dense geometry across diverse data, and some explicitly predict a disentangled metric scale factor.
Importantly, FF3R enables fast, optimization-free 3D estimation, offering a practical alternative to offline global reconstruction pipelines while producing geometry that is both more stable than monocular lifting and less coupled to a specific depth sensor.

Motivated by these observations, we propose \ourmodel, as shown in \cref{fig:teaser}, an \emph{online} multi-view 3D object detection framework that turns a FF3R model into the geometric encoder of a detection transformer~\cite{carion2020end,zhang2022dino,yang20253dmood}, and predicts 3D boxes directly in the reconstructed 3D space from RGB images.
This design aims to marry image-only systems with the geometric clarity of 3D reasoning, improving robustness to camera and domain shifts that often break the conventional 2D-to-3D lifting pipeline.
Concretely, we extend MapAnything~\cite{keetha2026mapanything} for its camera-conditioned modeling and disentangled metric-scale prediction, and repurpose it as the encoder that produces 3D tokens for a detection transformer (DETR) style detection head~\cite{carion2020end,zhang2022dino}.
On top of this backbone, we introduce an \emph{up-to-scale} 3D box head: it regresses boxes in an unscaled coordinate system and recovers metric outputs via the FF3R scale factor, thereby sidestepping the brittle depth regression inherent in 2D detections.
To satisfy online constraints, \ourmodel operates on a sliding temporal window and aggregates evidence causally with a multi-view transformer; camera intrinsics and, when available, poses provide explicit geometric cues that further stabilize depth and scale.

We focus on class-agnostic detection to isolate the geometric bottleneck and train \ourmodel on the CA-1M~\cite{lazarow2025cubify} dataset, but the architecture is not inherently class-specific and can be extended to semantic categories using the same geometric backbone.
\ourmodel achieves the new state-of-the-art (SOTA) performance on the held-out scenes of CA-1M.
Furthermore, we show the generalization ability of \ourmodel on the zero-shot ScanNetV2~\cite{dai2017scannet} benchmark.

To conclude, our contributions are threefold.
\textbf{(1)} We introduce \ourmodel, an RGB-only online multi-view indoor 3DOD framework that injects feed-forward metric reconstruction priors into a detection transformer to enable direct 3D reasoning from images.
\textbf{(2)} We propose an up-to-scale, direct 3D bounding-box head that recovers metric scale through FF3R, avoiding the error-amplifying 2D-to-3D lifting pathway.
\textbf{(3)} ~a detailed component analysis showing that the gains come from geometry-and-scale design choices (\ie direct 3D prediction, temporal multi-view aggregation, and explicit camera conditioning), together with strong generalization across indoor benchmarks.

%% file: sec/02_Related.tex
\section{Related Work}
\label{sec:related}

\subsection{Monocular 3D Object Detection}
Estimating metric 3D bounding boxes from a single RGB image is widely studied in robotics and autonomous driving.
Existing approaches span proposal-based pipelines, explicit geometric priors, and BEV-style formulations~\cite{hu2022monocular,park2021dd3d,liu2022petr,cc3dt,lin2022sparse4d,li2022bevformer,yang2023bevformer,tu2023imgeonet,naiden2019shift}.
A recent trend is to improve robustness across domains, cameras, and label spaces. Cube R-CNN~\cite{brazil2023omni3d} introduces virtual depth to mitigate focal-length variability when mixing datasets, while Uni-MODE~\cite{li2024unimode} uses domain confidence to jointly train BEV detectors across indoor and outdoor scenes.
Open-vocabulary variants increasingly leverage 2D foundation models: OVM3D-Det~\cite{huang2024training} generates pseudo supervision for novel classes, and 3D-MOOD conditions detection on vision--language priors (\eg GroundingDINO~\cite{liu2023grounding}) to reduce closed-set dependence.
Nevertheless, monocular 3DOD remains fundamentally constrained by single-view ambiguity: depth and, in particular, metric scale are underconstrained, so accurate geometry often relies on learned priors that can be sensitive to domain shifts in intrinsics, motion, and scene statistics~\cite{piccinelli2024unidepth,bochkovskii2024depthpro}.

\subsection{Feed-Forward 3D Estimation}
Feed-forward 3D estimation has improved substantially with large-scale pretraining, yielding single-view models with strong geometric representations and transfer across datasets and camera setups~\cite{piccinelli2025unik3d,piccinelli2025unidepthv2,hu2024metric3dv2,yang2024da2}.
Multi-view feed-forward estimators further reduce ambiguity by enforcing cross-view consistency and by jointly reasoning about camera motion and depth~\cite{wang2025vggt,lin2025da3,keetha2026mapanything}.
However, most of these approaches are scene-centric: they target camera motion and dense geometry, and are not optimized to produce object instances or object-centric representations.
In addition, many reconstruction pipelines assume global aggregation over a set of views, whereas streaming perception must update predictions causally from a short temporal context and within limited compute.

\subsection{Online 3D Perception}
Online 3D perception exploits temporal continuity to reduce ambiguity in streaming settings.
For geometry, prior works extend single-view predictors to video via propagation or temporal refinement~\cite{chen2025vda,piccinelli2025velodepth}, and others adapt multi-view reconstruction to incremental pipelines that update structure as frames arrive~\cite{zhuo2026streamvggt,duisterhof2025mastersfm}.
For object-centric reasoning, BoxFusion~\cite{lan2025boxfusion} combines pretrained 3D detectors with temporal persistence and post-processing, while EFM3D~\cite{straub2024efm3d} fuses multiple modalities to predict 3D occupancy.
Many online systems, however, either rely on additional sensors/modalities or retain a 2D-then-lift design, so it remains unclear how to obtain stable metric 3D boxes from RGB video alone under causal constraints.

\ourmodel addresses this gap by operating on a sliding window of frames, \ie treating time as additional views, repurposing MapAnything~\cite{keetha2026mapanything} as a geometry-and-scale encoder within a DETR-style detector, and decoding 3D boxes directly in 3D.
This avoids depth sensors and a 2D-to-3D lifting stage, while leveraging metric-scale priors learned through feed-forward multi-view reconstruction.

%% file: sec/03_Method.tex
\section{Method}
\label{sec:method}

\begin{figure}[t]
    \centering
    \footnotesize
    \includegraphics[width=1.0\linewidth]{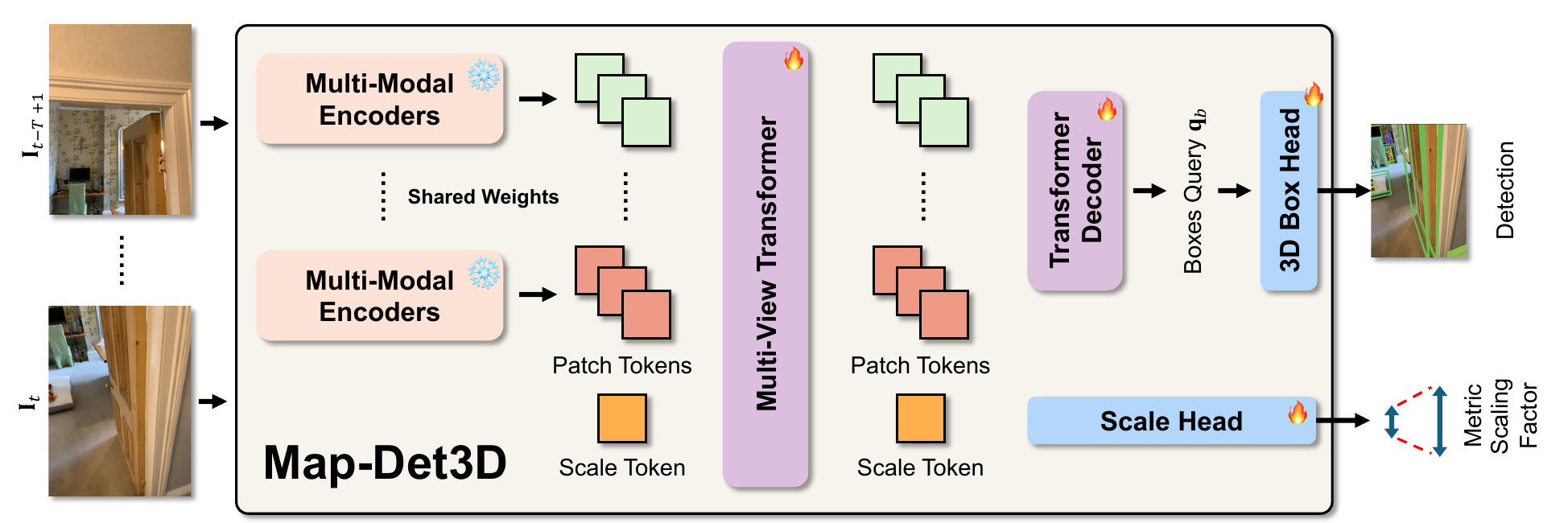}
    \caption{
        \textbf{Architecture overview.}
        \ourmodel treats a short temporal window as multi-view input, repurposes a metric FF3R backbone, \ie MapAnything, as the encoder of a detection transformer, and decodes up-to-scale 3D boxes that are converted to metric scale via the predicted scale factor.
    }
    \label{fig:overview}
    \vspace{5pt}
\end{figure}

Our goal is online 3D object detection in metric scale from streaming RGB, without depth sensors. One of the obstacles is the geometric component: from a single view, depth and absolute scale are underconstrained, and small range errors quickly dominate 3D overlap.
We address this by injecting a metric geometric prior into the detector.
Concretely, we repurpose a FF3R model as the geometric encoder of a detection transformer, and we introduce an up-to-scale 3D bounding box head whose outputs are mapped to metric units through a predicted scale factor.
\cref{fig:overview} summarizes the architecture.
We first formalize the online detection setup and notation (\cref{sec:method:preliminaries}), then describe the FF3R prior (\cref{sec:method:mapanything}), the detection architecture (\cref{sec:method:detection}), and the up-to-scale 3D head (\cref{sec:method:box3d_head}), and finally discuss online inference (\cref{sec:method:mv}) and training losses (\cref{sec:method:loss}).

\subsection{Preliminaries}
\label{sec:method:preliminaries}

We address the problem of online 3D object detection (3DOD) from a streaming monocular video $\video \coloneqq \{\img_1, \img_2, \ldots\}$, where we expect the model to produce metric-scale predictions causally, without depth sensors.
Each frame $\img_t \in \mathbb{R}^{3 \times H \times W}$ may be accompanied by camera intrinsics $\camK_t \in \mathbb{R}^{3 \times 3}$ and extrinsics $\camP_t \in \mathbb{R}^{4 \times 4}$.
When some of this metadata is unavailable, our FF3R backbone can provide pose and scale estimates, so the same formulation covers settings with complete, partial, or missing camera parameters.
At each time step $t$, the detector observes a sliding window $\window_t \coloneqq \{\img_{t-\winsize+1}, \ldots, \img_{t}\}$ of $\winsize$ frames. We denote by $\cam_t$ the set of camera parameters available for the window.
From $\window_t$ and $\cam_t$, the model outputs a fixed-size set of~$\nqueries$ 3D oriented bounding boxes $\detset_{t}$ for the current frame.
Each box in $\detset_{t}$ is parameterized by a 3D center $(x,y,z)$ in the camera coordinate frame, physical dimensions $(w,l,h)$, an orientation $R\!\in\!\mathrm{SO}(3)$, and a confidence score $\sigma\!\in\![0,1]$ that indicates whether it corresponds to a valid object.
A central element of our design is a per-window scale factor $\scale_t$, predicted by the backbone, that converts up-to-scale predictions (denoted with a tilde, \eg $\tilde{x}$) into metric quantities (\eg $x = \scale_t\,\tilde{x}$).
We focus on the class-agnostic setting to isolate the geometric challenges of depth and scale from category recognition.

\subsection{Metric Feed-Forward 3D Reconstruction Prior}
\label{sec:method:mapanything}
We build on MapAnything~\cite{keetha2026mapanything} as our metric FF3R backbone for two reasons: it produces a rich multi-view representation, and it predicts a disentangled per-window scale factor $\scale_t$.
Given the $\winsize$ frames in the current window $\window_t$ and the available camera parameters $\cam_t$, a multi-modal encoder produces per-view patch tokens that are then processed, together with a learnable scale token $\qscale$, by a 16-layer multi-view transformer~\cite{wang2025vggt} that fuses information across views.
We denote by $\featenc \in \mathbb{R}^{\winsize \times 1536 \times \npatches}$ the fused encoder output and by $\mathbf{F}_7, \mathbf{F}_{11}, \mathbf{F}_{15}$ the intermediate transformer features, each concatenated across all $\winsize$ views, where $\npatches \coloneqq HW/p^2$ is the number of spatial tokens per view and $p = 16$ is the patch size of the vision encoder.
These form the multi-scale feature maps $\msfeats_t \coloneqq \{\featenc, \mathbf{F}_7, \mathbf{F}_{11}, \mathbf{F}_{15}\}$.
An MLP then decodes the transformer-updated token corresponding to $\qscale$ into the scale factor $\scale_t$.
When ground-truth extrinsics are not included in $\cam_t$, the backbone estimates them internally.
In summary, the backbone mapping can be written as:
\begin{equation}
\label{eq:ff3r}
(\msfeats_t,\, \scale_t) \coloneqq \ffr(\window_t,\, \cam_t),
\end{equation}
and its output is used by subsequent modules in our architecture.
For brevity, we drop the time index $t$ from $\msfeats_t$, $\scale_t$, and all derived quantities in the following subsections, as the dependency is always through the input window.

\begin{figure}[t]
    \centering
    \footnotesize
    \includegraphics[width=0.9\linewidth]{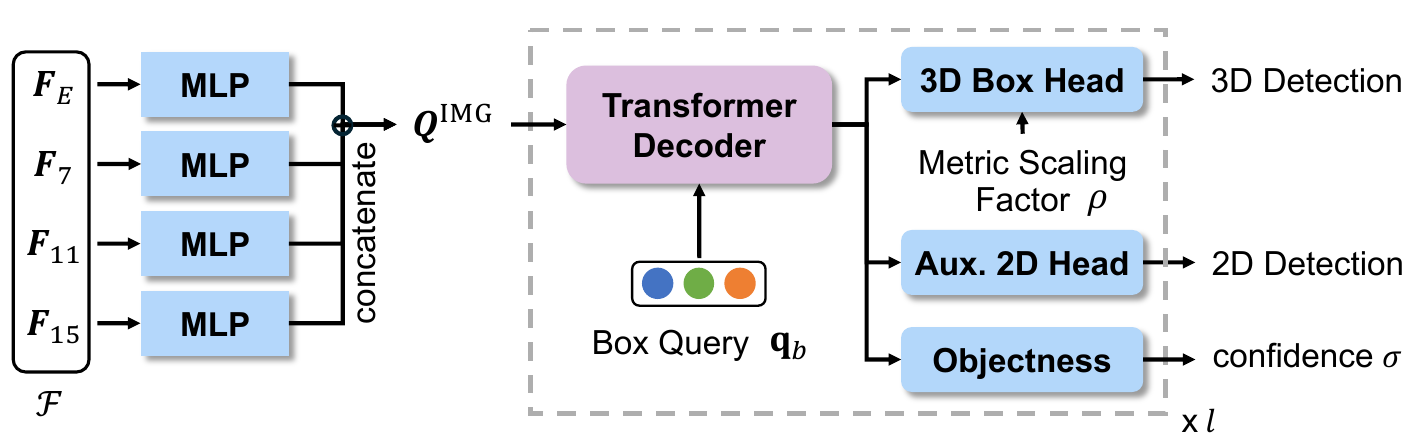}
    \caption{
        \textbf{Detection Architecture.}
        We repurpose the FF3R as the detection transformer~\cite{carion2020end,zhang2022dino} encoder and project the multi-scale feature maps into a common dimension, then concatenate them to form $\qimg$.
        We gradually refine bounding box queries and decode 3D bounding boxes using our proposed up-to-scale 3D bounding box head.
    }
    \label{fig:detection_arch}
\end{figure}

\subsection{Detection Architecture}
\label{sec:method:detection}
To transfer the FF3R geometric prior to 3DOD, we build a detection transformer~\cite{carion2020end,zhang2022dino,yang20253dmood} on top of the multi-scale feature maps $\msfeats$ as shown in \cref{fig:detection_arch}.
The $\winsize$ views are split apart, and each is processed independently as a separate batch element.
For each view, the feature levels are projected along the channel dimension to a common size of $256$ via a per-level MLP and concatenated along the patch and scale dimensions into a single token sequence, forming per-view image features $\qimg \in \mathbb{R}^{|\msfeats|\npatches \times 256}$.
Dense 2D anchor proposals are generated on a regular grid at each feature level with scale-dependent default sizes.
A shared classification and regression head scores all proposals. The top-$\nqueries$ scoring positions are selected as initial reference points, and the corresponding encoder features, after stop-gradient and linear projection, form the initial object queries $\qbox_{0} \in \mathbb{R}^{\nqueries \times 256}$.
This top-$\nqueries$ selection determines the fixed output size introduced in \cref{sec:method:preliminaries}.

A $\nlayers$-layer deformable decoder~\cite{zhang2022dino} then iteratively refines the queries. Each layer applies deformable cross-attention to $\qimg$, self-attention among queries, and updates the reference points via residual 2D box refinement.
We denote this encoder-decoder pipeline as $\detmod$, which outputs the queries $\qbox_{k}$ and the associated 2D reference boxes $\detboxtwoD_{k}$ at every layer $k$:
\begin{equation}
\{(\qbox_{k},\, \detboxtwoD_{k})\}_{k=0}^{\nlayers} \coloneqq \detmod(\msfeats).
\end{equation}
Each $\qbox_{k}$ is decoded into an intermediate 3D box prediction (\cref{sec:method:box3d_head}), allowing progressive refinement across layers, while the 2D boxes $\detboxtwoD_{k}$ provide the reference points for deformable attention and are used for bipartite matching during training (\cref{sec:method:loss}).

\subsection{Up-to-scale 3D Bounding Box Head}
\label{sec:method:box3d_head}
As shown in \cref{fig:detection_arch}, at every decoder layer $k$, we apply a layer-specific 3D bounding box head $\headthreeD^{k}$ to the $\nqueries$ queries $\qbox_{k}$ to produce a set of $\nqueries$ 3D bounding boxes with associated confidence scores:
\begin{equation}
(\detset_{k},\, \detscores_{k}) \coloneqq \headthreeD^{k}(\qbox_{k},\, \scale), \quad k = 0, \ldots, \nlayers.
\end{equation}
The head receives the scale factor $\scale$ because all geometric attributes are regressed in an up-to-scale parameterization and must be converted to metric outputs.
The intermediate predictions $\detset_{0}, \ldots, \detset_{\nlayers-1}$ enable deep supervision during training (\cref{sec:method:loss}), while the final output of the model is $\detset_{t} \coloneqq \detset_{\nlayers}$ (reinstating the time index from \cref{sec:method:preliminaries}).
As introduced in \cref{sec:method:preliminaries}, each predicted box in $\detset_{k}$ is characterized by a 3D center $(x,y,z)$ in the camera coordinate frame, physical dimensions $(w,l,h)$, an orientation $R\!\in\!\mathrm{SO}(3)$, and a confidence score $\sigma\!\in\![0,1]$.
A dedicated two-layer MLP regresses each geometric attribute in an up-to-scale parameterization, which is then converted to metric scale via the scale factor $\scale$.

For the 3D center, the head regresses up-to-scale coordinates $\tilde{x}$, $\tilde{y}$ in camera space and a log-depth $\tilde{d}$, from which the metric center is recovered as $x \coloneqq \scale\,\tilde{x}$, $y \coloneqq \scale\,\tilde{y}$, and $z \coloneqq \scale\,\exp(\tilde{d})$.
Similarly, the head regresses up-to-scale log-dimensions $\tilde{s}_w$, $\tilde{s}_l$, $\tilde{s}_h$, from which the metric sizes are recovered as $w \coloneqq \scale\,\exp(\tilde{s}_w)$, $l \coloneqq \scale\,\exp(\tilde{s}_l)$, and $h \coloneqq \scale\,\exp(\tilde{s}_h)$.
For orientation, following Cube R-CNN~\cite{brazil2023omni3d}, we regress a continuous 6D rotation representation~\cite{zhou2019continuity} in allocentric form and convert it to egocentric using the predicted 3D center direction~\cite{kundu20183d}, unlike CuTR~\cite{lazarow2025cubify}, which assumes all bounding boxes are gravity-aligned.
Finally, a single-layer MLP predicts a binary objectness logit yielding the confidence score $\detscores$.
Each query also carries the 2D reference box $\detboxtwoD_{k}$ from the decoder, which is used for bipartite matching during training (\cref{sec:method:loss}).

\subsection{Online Setting}
\label{sec:method:mv}

\PAR{Inference.} \ourmodel operates causally with a sliding window. At time $t$, it uses the current frame and the previous $\winsize{-}1$ frames as multi-view input and outputs detections only for the current frame.
This reuses MapAnything's multi-view fusion for temporal reasoning without modifying the backbone's architecture.

\PAR{Training.} Following VGGT~\cite{wang2025vggt}, we randomize the number of views by sampling uniformly from $1$ to the maximum window size $\winsize$, which improves robustness to varying temporal context.

\subsection{Training Losses}
\label{sec:method:loss}
We combine auxiliary 2D losses for matching with disentangled 3D losses for geometry.
We compute Hungarian matching~\cite{kuhn1955hungarian} based on the auxiliary 2D box predictions, and supervise matched pairs with focal loss~\cite{lin2017focal} for classification and $L_1$+GIoU~\cite{rezatofighi2019generalized} for 2D box regression.
For 3D geometry, we use a disentangled corner loss as in Cube R-CNN~\cite{brazil2023omni3d}.
For each attribute, we form the eight 3D box corners using the prediction for that attribute and ground truth for all others.
This yields $\Lxy$ (center), $\Lz$ (depth), $\Ldim$ (dimensions), and $\Lrot$ (rotation), where $\Lrot$ is a Chamfer distance between corner sets to handle symmetries.
The remaining terms use $L_1$.
We apply deep supervision to the encoder proposal layer and all decoder layers, giving the total per-frame loss:
\begin{equation}
\Ltotal \coloneqq \sum_{k=0}^{\nlayers} ( \LtwoD^{k} + \LthreeD^{k} ), \quad \LthreeD^{k} \coloneqq \Lxy^{k} + \Lz^{k} + \Ldim^{k} + \Lrot^{k}.
\end{equation}
The final training objective is the expectation of $\Ltotal$ over all sampled frames.

%% file: sec/04_Experiments.tex
\section{Experiments}
\label{sec:exp}

We first discuss the implementation details for model training and testing in \cref{sec:exp:implement}.
We then describe the experimental setup in \cref{sec:exp:setup}, including the training and testing data and the evaluation metrics.
In \cref{sec:exp:analysis}, we take a deep look into \ourmodel through extensive ablation studies, and in \cref{sec:exp:benchmark}, we compare \ourmodel with other state-of-the-art (SOTA) models on both in-domain and out-of-domain benchmarks. 
In \cref{sec:exp:per_scene}, we extend \ourmodel with a simple 3D object tracking algorithm to directly compare our method on an online per-scene evaluation benchmark.
Finally, we provide the qualitative results in \cref{sec:exp:qualitative_comp}.

\subsection{Implementation Details}
\label{sec:exp:implement}
We implement \ourmodel in PyTorch~\cite{pytorch} and CUDA~\cite{nickolls2008cuda}.
We train the full model for $100k$ steps with a batch size of $64$, \ie, a maximum of $4$ samples per GPU on $16$ RTX 4090s for $1.5$ days.
For ablation studies in \cref{sec:exp:analysis}, we train the model for $50k$ steps, also with a batch size of $64$.
We set the initial learning rate to $0.0001$ and apply the Cosine Annealing learning rate scheduler.
To fully leverage the multi-view transformer for temporal information for 3DOD, we unfreeze the multi-view transformer and the scale head by setting their learning rate to $1/10$ of the initial learning rate.
For input image resolutions, we follow MapAnything by selecting the closest image ratio from the fixed mapping.
Instead of resizing the long edge of the image and performing center cropping, we resize the short edge and pad with zeros to preserve global information.
During training, we randomly select the number of views from $1$ to $5$ and the aspect ratio, and randomly select the sampling rate from $2$ to $10$ FPS as part of the data augmentation.
During inference time, we set $\nviews=5$.

\subsection{Experimental Setup}
\label{sec:exp:setup}
\PAR{CA1M~\cite{lazarow2025cubify}.}
We use the dataset to train \ourmodel and evaluate on validation for an in-domain benchmark.
Unlike previous datasets that focus on a limited set of predefined object categories, CA-1M provides exhaustive, class-agnostic 9-DOF 3D bounding box annotations.
It features over $400,000$ unique 3D objects across more than $1,000$ highly accurate, laser-scanned indoor scenes.
These high-fidelity 3D annotations are registered to over $3,500$ handheld, egocentric video captures, yielding approximately $13$ million training frames and $1.8$ million validation frames.
The pixel-accurate alignment of metric 3D boxes across continuous video sequences makes CA-1M a good fit for training \ourmodel to learn consistent metric scale and temporal stability under causal, moving-camera conditions.

\PAR{ScanNetV2~\cite{dai2017scannet}.}
We further validate the generalization and robustness of our proposed architecture, given its widespread adoption as a benchmark.
ScanNetV2 contains $1,513$ densely annotated indoor scenes standardized into $1,201$ scenes for training and $312$ for validation.
The dataset provides continuous RGB video streams paired with ground-truth 3D bounding box annotations with $18$ core (ScanNetV2) and $198$ long-tailed semantic categories (ScanNet200).
We follow BoxFusion~\cite{lan2025boxfusion} to use the selected $100$ scenes and uniformly sample every $25$ frames as the zero-shot benchmark.
We use the challenging ScanNet200 settings for per-frame 3DOD evaluation and the original ScanNetV2, \ie $18$ classes, as the per-scene evaluation to be aligned with BoxFusion.

\PAR{Evaluation Metrics.}
We use the average precision (AP) and average recall (AR) metrics to evaluate the performance of 2D and 3D detection results.
We match the predictions and GT by computing the intersection-over-union ($\text{IoU}_{\text{3D}}$) of 3D cuboids.
The mean 3D AP is reported in a class-agnostic way, while $\text{AP}_{15}$, $\text{AP}_{25}$, and $\text{AP}_{50}$ correspond to $\text{IoU}_{\text{3D}}$ thresholds $0.15$, $0.25$ and $0.50$, respectively.
It is worth noting that, unlike in previous benchmark~\cite{brazil2023omni3d,yang20253dmood}, we evaluate \textit{all} ground truth bounding boxes regardless of their visibility and truncation.

\input{tab/ablation}

\subsection{Model Analysis}
\label{sec:exp:analysis}

We first provide a detailed analysis of \ourmodel in \cref{tab:results:ablation} and \cref{tab:rebuttal:direct_3d} to assess the effectiveness of our model designs.
Then, we analyze the efficiency in \cref{tab:rebuttal:fps}.

\parsection{Feed-Forward 3D Reconstruction Prior.}
We first ablate the different settings for leveraging the FF3R prior.
In particular, we analyze the exploitation of the multi-view prior \vs single-view and whether the prior components, \ie scale prediction and multi-view feature aggregator, are either frozen or trainable.
Enabling multi-view processing without further adapting the FF3R prior yields only a negligible gain (row~1 $\rightarrow$ row~2), which indicates that directly using the multi-view transformer to aggregate temporal information for 3DOD is not straightforward.
Unfreezing the scale head alone also has a limited impact (row~2 $\rightarrow$ row~3), indicating that the original scale prediction prior does not require further fine-tuning on CA-1M.

In contrast, fine-tuning the multi-view transformer is important for detection: unfreezing it leads to a clear improvement (row~3 $\rightarrow$ row~4).
Under the same fine-tuning setup, adding multi-view input provides further gains (row~4 $\rightarrow$ row~5), supporting the role of temporal context beyond parameter updates.
These clear improvements highlight that, although the geometric prior provides strong 3D information, it lacks sufficient object-centric features that must be trained to yield the largest gain.

\parsection{Camera conditioning.}
We further show how \ourmodel can benefit from camera conditioning on FF3R.
As shown in row~2 $\rightarrow$ row~6 of \cref{tab:results:ablation}, conditioning the camera intrinsics to MapAnything's multi-modal encoder moderately improves performance by $0.8~\text{AP}_\text{15}$.
Adding camera poses yields a further substantial boost of $4.7~\text{AP}_\text{15}$ (row~6 $\rightarrow$ row~7), which shows that the \ourmodel can successfully leverage the FF3R prior and perform better 3DOD accordingly.

Finally, enabling all components and training the full model achieves the best result, with a considerable gap of $9.5~\text{AP}_\text{15}$ \wrt the baseline (row~1 $\rightarrow$ row~8).
The final model shows how our design contributed with a different set of complementary contributions: metric-scale priors, temporal aggregation, and explicit camera cues.

\parsection{3D Bounding Box Head.}
By repurposing MapAnything as the detection transformer encoder, \ourmodel can leverage the strong geometric prior, thereby bypassing the 2D-to-3D lifting from 2D pixel-space detection.
\cref{tab:rebuttal:direct_3d} confirm its effectiveness by improving both $\text{AP}_\text{15}$ for both in-domain and out-of-domain settings.

\PAR{Efficiency.} We provide the temporal ablation along with the runtime and cost in \cref{tab:rebuttal:fps}, which shows that \ourmodel benefits from the temporal contexts.
Given the maximum $T=5$ during training, testing with $T=7$ leads to a minor degradation ($-0.1$ AP$_\text{15}$) on CA1M (in-domain), yet we attain the same performance on ScanNet (out-of-domain).
We further conduct the few-view setting (last row), \ie strides = $5$, which obtains $T$-frame detection results at once.
This offline setting leads to slightly better results, while our online sliding-window design shows competitive performance.

\input{tab/rebuttal}

\input{tab/ca1m}

\subsection{State-of-the-art Comparisons}
\label{sec:exp:benchmark}
\parsection{In-Domain Benchmark.}
\cref{tab:results:ca1m} reports results on CA-1M validation sets, \ie the scenes are not seen during training.
\ourmodel achieves $16.9~\text{AP}_{25}$ and $3.5~\text{AP}_{50}$, improving over monocular baselines such as CuTR~\cite{lazarow2025cubify} ($13.5$ / $2.4$) and Cube R-CNN~\cite{brazil2023omni3d} ($4.6$ / $1.0$).
Furthermore, \ourmodel also surpasses the offline multi-view baseline ImVoxelNet~\cite{rukhovich2022imvoxelnet} ($10.1$ / $2.3$) while operating online with a sliding window.
This demonstrates the effectiveness of our model design in leveraging multi-view information via FF3R's priors.
Despite depth/point-cloud methods remaining stronger (\eg FCAF~\cite{rukhovich2022fcaf3d}: $29.3$ / $11.2$), as expected given their access to metric measurements, \ourmodel closes the gap \wrt TR3D ($22.0$ / $4.4$), demonstrating the strengths of our designs.

\input{tab/scannet}

\parsection{Zero-shot benchmark.}
In \cref{tab:results:scannet}, we evaluate all methods under the class-agnostic setting on the ScanNet200 validation set without fine-tuning.
Relative to open-set monocular baselines trained on other sources, \ourmodel achieves the best zero-shot performance on AP$_{15}$ and AP$_{25}$.
This demonstrates the generalization ability even though the model is trained only on a single source dataset, \ie CA-1M.
We further test CuTR, the only other 3D object detector trained on CA-1M to our best knowledge, on ScanNetV2.
While \ourmodel reaches $15.2~\text{AP}_{15}$ and $9.7~\text{AP}_{25}$, CuTR only achieve $4.3~\text{AP}_{15}$.
This indicates that \ourmodel's superior performance is not obtained by CA-1M training data but rather by our model design.

\subsection{Per-scene Evaluation}
\label{sec:exp:per_scene}

As we conduct per-frame evaluation in \cref{tab:results:ca1m} and \cref{tab:results:scannet}, we also follow the setting in BoxFusion~\cite{lan2025boxfusion} to have per-scene results.
To achieve this, instead of using the optimization process proposed in BoxFusion, we follow the widely used tracking-by-detection paradigm~\cite{hu2022monocular,cc3dt} and track the 3D detection results frame by frame in an online manner.

\begin{figure}[t]
    \centering
    \footnotesize
    \includegraphics[width=1.0\linewidth]{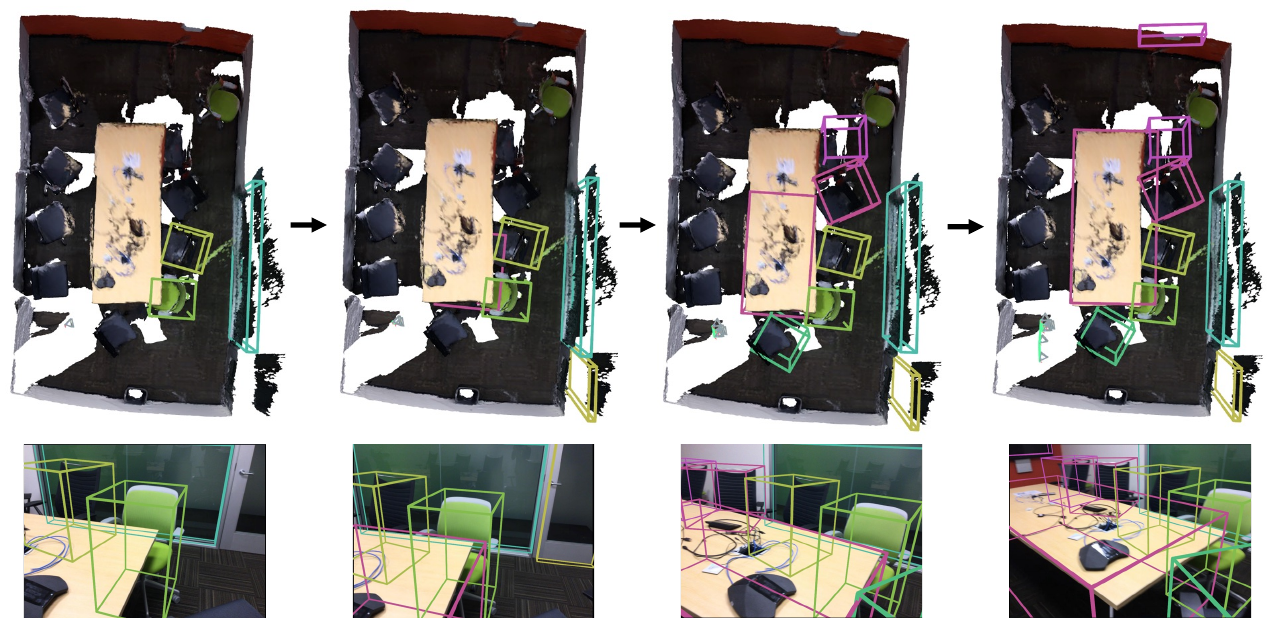}
    \vspace{-5pt}
    \caption{
        \textbf{The effect of the visible part inside the images.}
        The table is only partially observed when T = 1.
        But as the camera rotates, the model can gradually see the full table.
        Hence, our tracking design can successfully use the better detection results from \ourmodel as the per-scene results.
    }
    \label{fig:sup:seq}
\end{figure}

We maintain an \textit{online} track memory $\mathcal{T}$, and for each track $\tau \in \mathcal{T}$, it contains the detected 3D object in its corresponding camera coordinate and the camera pose.
For each new frame detection, \ie observation, we first transform both the observation and the detected boxes in track memory to world coordinates, and then compute the 3D IoU between them as the affinity matrix.
Then we will run the greedy assignment to determine whether the new detection should be assigned to a new track or associated with an existing one.
Finally, we will replace the bounding box of the associated track with the new observation's bounding box if it is larger.
As shown in \cref{fig:sup:seq}, we speculate that when two objects are associated in 3D space over time, the larger box usually corresponds to the more visible part in the image, and we update the track accordingly.

We report the zero-shot per-scene evaluation results on ScanNetV2 in \cref{tab:results:per_scene}.
\ourmodel with simple 3D IoU based association yields better performance than the previous one that did not use ground truth depth methods.
Moreover, we achieve better results on AP$_{15}$ than EmbodiedSAM and OnlineAnySeg when they uses GT depth information.
For stricter evaluation settings, \ie AP$_{25}$ and AP$_{50}$, we still have comparable performance against point-cloud-based methods.
This demonstrates that \ourmodel successfully leverages the FF3R prior and achieves new state-of-the-art (SOTA) performance on both per-frame and per-scene evaluations.

\input{tab/scene}

\input{fig/qualitative_comparison}

\subsection{Qualitative Results}
\label{sec:exp:qualitative_comp}
In \cref{fig:qualitative_comp}, we qualitatively compare \ourmodel with several baselines on the ScanNetV2 validation set (out-of-domain).
We overlay predicted 3D bounding boxes and visualize them together with the ground-truth scene geometry (shown as a point cloud) to assess box placement, scale, and orientation in 3D.
On CA-1M, CuTR commonly exhibits monocular failure modes such as depth offsets and scale drift, which lead to boxes that are visibly too large/small or shifted along the viewing direction.
\ourmodel produces boxes that better match the scene geometry in these cases, consistent with using a sliding-window view aggregation and an explicit metric scale factor from the FF3R prior.
Under zero-shot transfer to ScanNetV2, CuTR’s predictions degrade further, with larger scale and depth errors, whereas \ourmodel typically preserves more coherent metric scale and maintains plausible 3D alignment under domain, \eg appearance and camera shift.
We provide additional qualitative results on ScanNetV2 in \cref{fig:qualitative_scannet} and more in the supplementary material.

\input{fig/qualitative_scannet}

%% file: tab/ablation.tex
\begin{table}[t]
    \small
    \footnotesize
    \centering
    \setlength\tabcolsep{7.5pt}
    \caption{
        \textbf{Model Analysis.}
        We conduct extensive ablation studies on \ourmodel to analyze the effectiveness of each proposed design on the CA-1M validation set.
        \textbf{Multi-view} stands for using online temporal information, \textbf{MV Transformer} stands for multi-view transformer in MapAnything.
        \textbf{K} and \textbf{Pose} denote whether encoding the camera intrinsics and poses in the multi-modal encoder, respectively.
    }
    \vspace{-5pt}
    \resizebox{1\linewidth}{!}{
    \begin{tabular}{lc|cc|cc|c}
        \toprule
        & Multi-View & Unfreeze Scale Head & Unfreeze MV Transformer & K & Pose & $\text{AP}_\text{15}$ \\
        \midrule
        1 & - & - & - & - & - & 11.7 \\
        2 & \checkmark & - & - & - & - & 11.8 \\
        3 & \checkmark & \checkmark & - & - & - & 11.9 \\
        \midrule
        4  & - & \checkmark & \checkmark & - & - & 14.5 \\
        5 & \checkmark & \checkmark & \checkmark & - & - & 17.2 \\
        \midrule
        6 & \checkmark & - & - & \checkmark & - & 12.6 \\
        7 & \checkmark & - & - & \checkmark & \checkmark & 17.3 \\
        \midrule
        8  & \checkmark & \checkmark & \checkmark & \checkmark& \checkmark & \textbf{21.2} \\
        \bottomrule
    \end{tabular}
    }
    \vspace{5pt}
    \label{tab:results:ablation}
\end{table}

%% file: tab/rebuttal.tex
\begin{table}[t]
    \small
    \footnotesize
    \centering
    \begin{minipage}[t]{0.37\linewidth}
        \setlength\tabcolsep{10pt}
        \caption{
            \textbf{3D head ablation.}
            We ablate the proposed 3D head under the multi-view setting while the scale head and multi-view transformer are frozen.
        }
        \vspace{-7pt}
        \resizebox{1\linewidth}{!}{
        \begin{tabular}{c|cc}
            \toprule
            Direct 3D & $\text{AP}^\text{CA1M}_\text{15}$ & $\text{AP}^\text{ScanNet}_\text{15}$ \\
            \midrule
            - & 15.6 & 10.8 \\
            \checkmark & \textbf{17.3} & \textbf{12.6} \\
            \bottomrule
        \end{tabular}
        }
        \label{tab:rebuttal:direct_3d}
    \end{minipage}
    \hfill
    \begin{minipage}[t]{0.59\linewidth}
        \setlength\tabcolsep{10pt}
        \caption{
            \textbf{FPS and peak GPU memory.}
            $^*$ stands few-view settings (offline), \ie strides = $5$.
        }
        \vspace{-5pt}
        \resizebox{1\linewidth}{!}{
        \begin{tabular}{l|cc|cc}
            \toprule
            $T$ & GPU Mem (G) & FPS & $\text{AP}^\text{CA1M}_\text{15}$ & $\text{AP}^\text{ScanNet}_\text{15}$ \\
            \midrule
            1 & 5.8 & 14.3 & 14.0 & 11.1 \\
            3 & 6.1 & 11.1 & 20.5 & 13.1 \\
            5 & 6.4 & 8.3 & \textit{21.2} & \textit{13.3} \\
            7 & 6.7 & 6.3 & 21.1 & \textit{13.3} \\
            \midrule
            5$^*$ & 6.4 & 5.8 & \textbf{22.2} & \textbf{13.5} \\
            \bottomrule
        \end{tabular}
        }
    \label{tab:rebuttal:fps}
    \end{minipage}
\end{table}

%% file: tab/ca1m.tex
\begin{table}[t]
    \small
    \footnotesize
    \centering
    \setlength\tabcolsep{15pt}
    \caption{
        \textbf{Results on CA-1M~\cite{lazarow2025cubify}.}
        We compare \ourmodel against different detector types on the challenging CA-1M validation set.
        The results show that our design can fully leverage the FF3R prior and not just outperform all monocular baselines but also outperform an offline multi-view detector.
    }
    \vspace{-5pt}
    \resizebox{1\linewidth}{!}{
    \begin{tabular}{l|c|cccc}
        \toprule
        \textbf{Method} & Detector Type & $\text{AP}_{\text{25}} \uparrow$ & $\text{AR}_{\text{25}} \uparrow$ & $\text{AP}_{\text{50}} \uparrow$ & $\text{AR}_{\text{50}} \uparrow$ \\
        \midrule
        FCAF~\cite{rukhovich2022fcaf3d} & Point Cloud & 29.3 & 49.5 & 11.2 & 22.6 \\
        TR3D~\cite{rukhovich2023tr3d} & Point Cloud & 22.0 & 49.5 & 4.4 & 20.0 \\
        TR3D + FF~\cite{rukhovich2023tr3d} & Point Cloud & 24.8 & 52.9 & 4.7 & 21.0 \\
        \midrule
        ImVoxelNet~\cite{rukhovich2022imvoxelnet} & Multi-View & 10.1 & 22.8 & 2.3 & 6.3 \\
        \midrule
        Cube R-CNN~\cite{brazil2023omni3d} & Monocular & 4.6 & 20.1 & 1.0 & 4.7 \\
        CuTR~\cite{lazarow2025cubify} & Monocular & 13.5 & 35.4 & 2.4 & 12.9 \\
        \midrule
        \textbf{\ourmodel (Ours)} & Online Multi-view & \textbf{16.9} & \textbf{40.3} & \textbf{3.5} & \textbf{14.8} \\
        \bottomrule
    \end{tabular}
    }
    \label{tab:results:ca1m}
    \vspace{5pt}
\end{table}

%% file: tab/scannet.tex
\begin{table}[t]
    \small
    \footnotesize
    \centering
    \setlength\tabcolsep{15pt}
    \caption{
        \textbf{Results on ScanNet200~\cite{dai2017scannet}.}
        We test the generalization ability of \ourmodel compared to other monocular open-vocabulary 3DOD methods~\cite{yao2024open,zhang2025detect,yang20253dmood,yao2026labelany3d}.
        The results demonstrate that, despite being trained on less diverse data, \ourmodel still outperforms all other methods.
        The comparison with CuTR further demonstrates that the improved performance is not due to the CA-1M training data, but rather to our model design.
    }
    \vspace{-5pt}
    \resizebox{0.9\linewidth}{!}{
    \begin{tabular}{l|c|cc}
        \toprule
        \textbf{Method} & Training Data & $\text{AP}_{\text{15}} \uparrow$ & $\text{AP}_{\text{25}} \uparrow$ \\
        \midrule
        OVMono3D~\cite{yao2024open} & Omni3D~\cite{brazil2023omni3d} & 11.7 & 5.7 \\
        LabelAny3D~\cite{yao2026labelany3d} & MS-COCO 3D~\cite{yao2026labelany3d} & 11.2 & 4.9 \\
        3D-MOOD~\cite{yang20253dmood} & Omni3D~\cite{brazil2023omni3d} & 11.2 & 8.0 \\
        DetAny3D~\cite{zhang2025detect} & DA3D~\cite{zhang2025detect} & 11.7 & 8.0 \\
        \midrule
        CuTR~\cite{lazarow2025cubify} & \multirow{2}{*}{CA-1M~\cite{lazarow2025cubify}} & 4.3 & 2.1 \\
        \textbf{\ourmodel (Ours)} & & \textbf{15.2} & \textbf{9.7} \\
        \bottomrule
    \end{tabular}
    }
    \vspace{5pt}
    \label{tab:results:scannet}
\end{table}

%% file: tab/scene.tex
\begin{table}[t]
    \small
    \footnotesize
    \centering
    \setlength\tabcolsep{10pt}
    \caption{
        \textbf{Zero-shot class-agnostic per-scene evaluation on ScanNetV2.}
        We follow BoxFusion~\cite{lan2025boxfusion} and run \ourmodel with a simple 3D IoU based association algorithm.
        Our method achieves new SOTA results without using ground-truth depth.
    }
    \vspace{-5pt}
    \resizebox{1\linewidth}{!}{
    \begin{tabular}{l|cc|ccc}
        \toprule
        Method & Online & No GT Depth & $\text{AP}_{\text{15}} \uparrow$ & $\text{AP}_{\text{25}} \uparrow$ & $\text{AP}_{\text{50}} \uparrow$ \\
        \midrule
        EmbodiedSAM~\cite{xu2024esam} & - & - & 8.2 & 5.2 & 1.4 \\
        
        OnlineAnySeg~\cite{tang2025onlineanyseg} & \checkmark & - & 24.2 & 18.3 & 5.2 \\
        BoxFusion~\cite{lan2025boxfusion} & \checkmark & - & 29.2 & 24.6 & 8.0 \\
        \midrule
        SpatialLM~\cite{SpatialLM} & - & \checkmark & 9.0 & 4.9 & 0.6 \\
        BoxFusion (RGB Only)~\cite{lan2025boxfusion} & \checkmark & \checkmark & 18.5 & 11.8 & 1.0 \\
        \textbf{Ours} & \checkmark & \checkmark & \textbf{27.6} & \textbf{22.7} & \textbf{6.2} \\
        \bottomrule
    \end{tabular}
    }
    \label{tab:results:per_scene}
\end{table}

%% file: fig/qualitative_comparison.tex
\begin{figure}[t]
    \centering
    \small
    \footnotesize
    \setlength{\tabcolsep}{1.5pt}
    \newcommand{\sz}{0.24}
    \begin{tabular}{ccc}
        \includegraphics[width=0.24\linewidth]{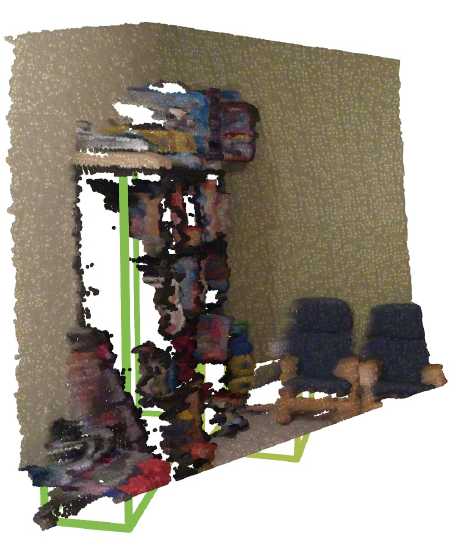} &
        \includegraphics[width=0.26\linewidth]{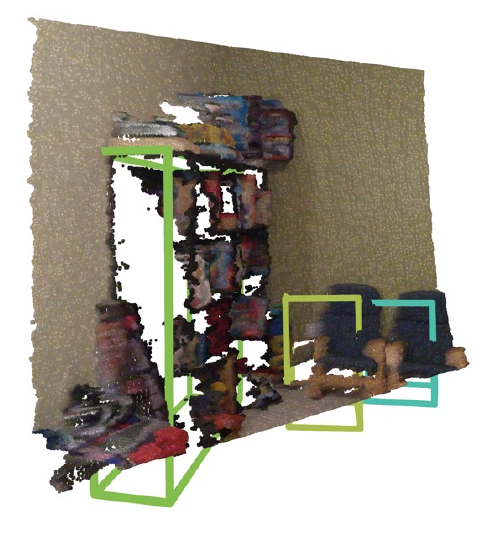} &
        \includegraphics[width=0.27\linewidth]{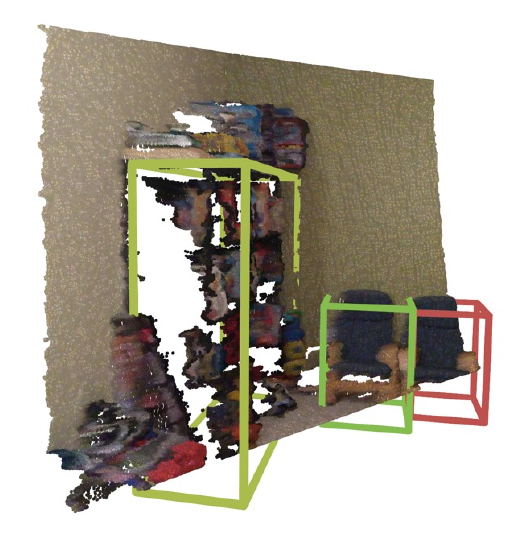} \\ 
        3D-MOOD & DetAny3D & \textbf{Ours} \\
    \end{tabular}
    \begin{tabular}{cccc}
        \includegraphics[width=\sz\linewidth]{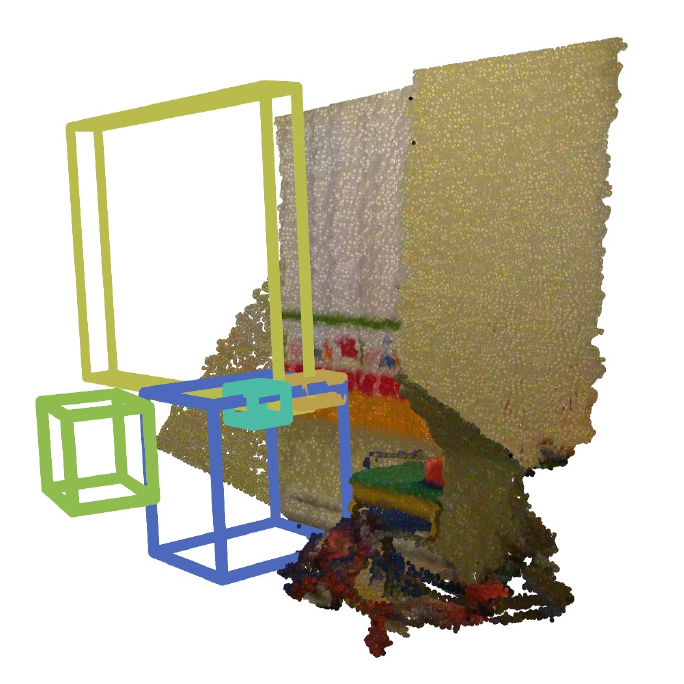} &
        \includegraphics[width=\sz\linewidth]{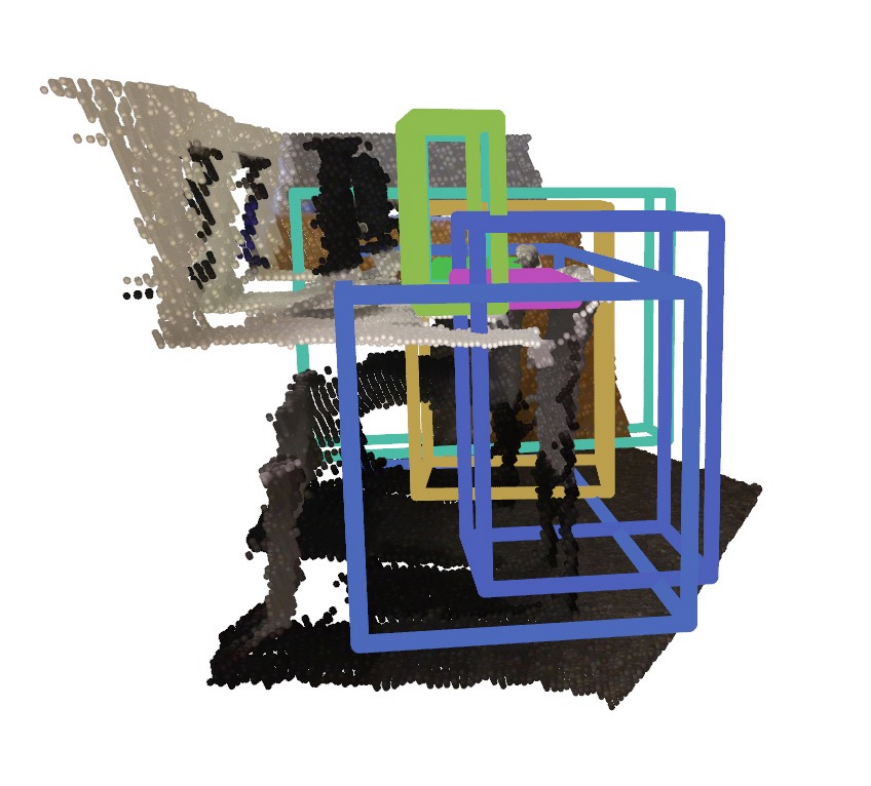} &
        \includegraphics[width=\sz\linewidth]{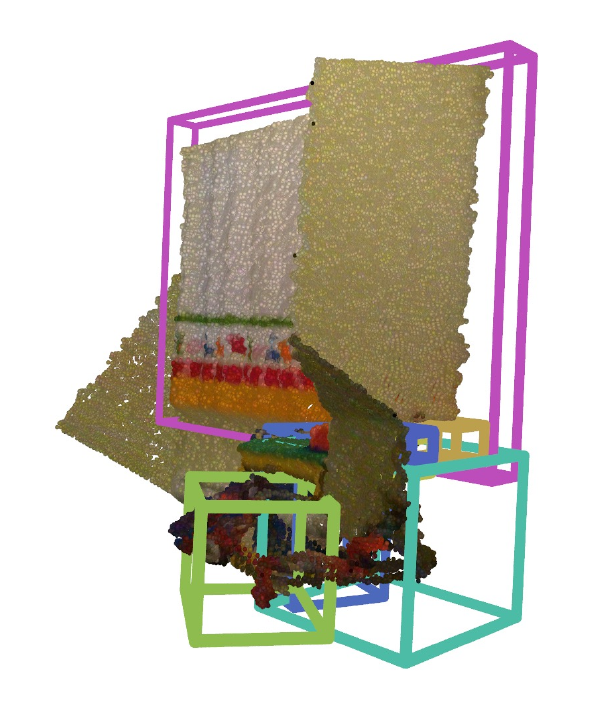} &
        \includegraphics[width=\sz\linewidth]{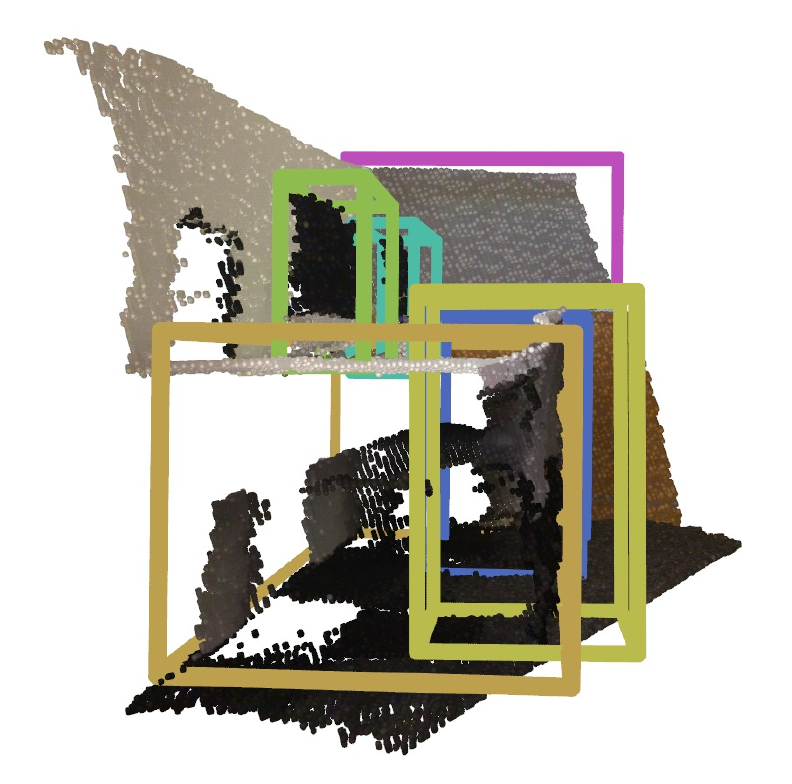} \\
        \multicolumn{2}{c}{CuTR~\cite{lazarow2025cubify}} & \multicolumn{2}{c}{\textbf{Ours}} \\
    \end{tabular}
    \caption{
        \textbf{Qualitative comparison on ScanNet.}
        We qualitatively compare \ourmodel with, 3D-MOOD~\cite{yang20253dmood}, DetAny3D~\cite{zhang2025detect}, and CuTR~\cite{lazarow2025cubify}.
        We visualize the predicted 3D bounding boxes alongside the ground-truth point cloud to qualitatively assess the accuracy of metric-scale prediction.
        The results demonstrate that, compared to the baselines, \ourmodel successfully detects objects given unseen scenes.
    }
    \label{fig:qualitative_comp}
\end{figure}

%% file: fig/qualitative_scannet.tex
\begin{figure}[t]
    \centering
    \small
    \footnotesize
    \setlength{\tabcolsep}{1.5pt}
    \newcommand{\szi}{0.24}
    \newcommand{\szo}{0.25}
    \newcommand{\hs}{5cm}
    \begin{tabular}{cccc}
    \includegraphics[width=\szi\linewidth]{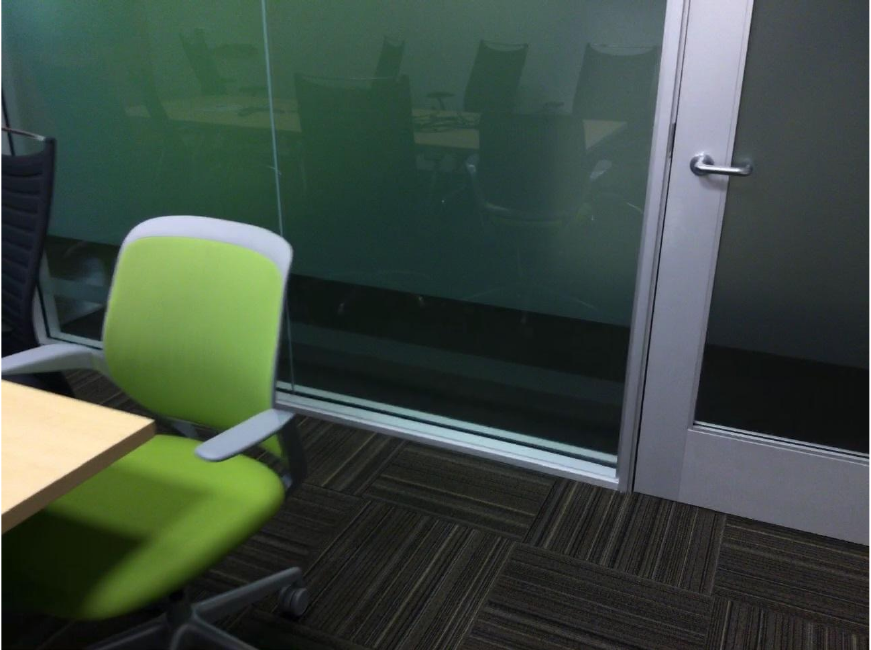} & \includegraphics[width=\szi\linewidth]{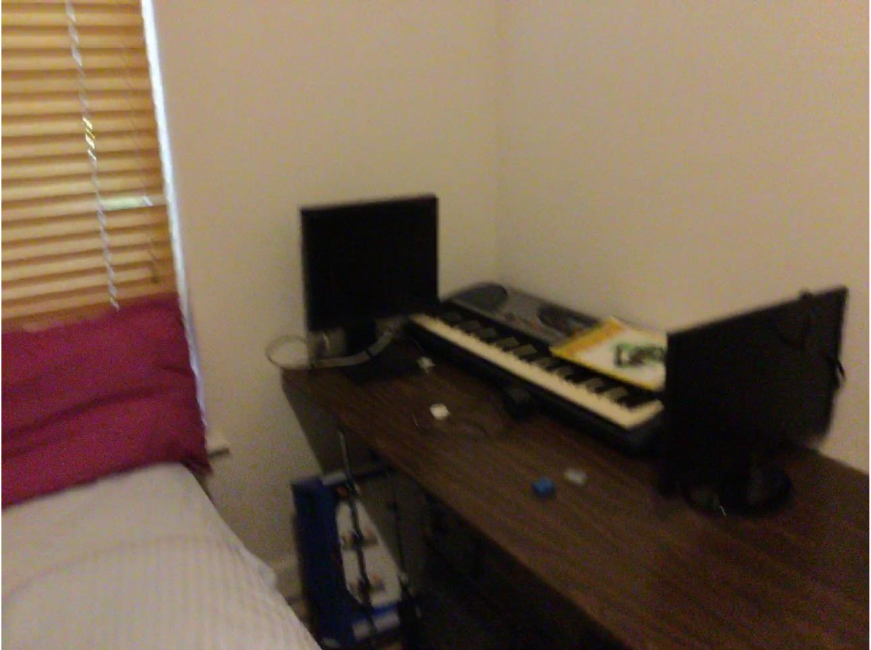} & \includegraphics[width=\szo\linewidth]{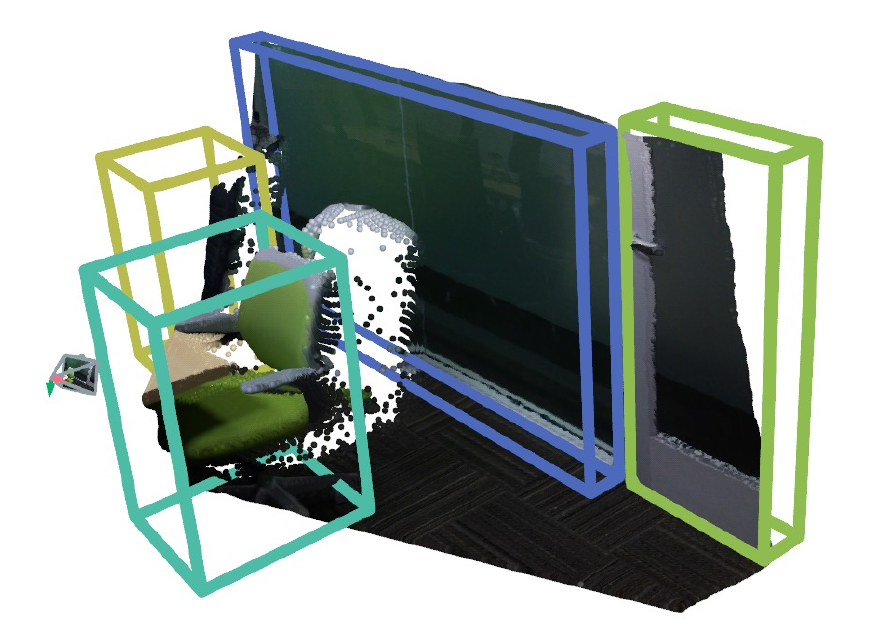} & \includegraphics[width=\szo\linewidth]{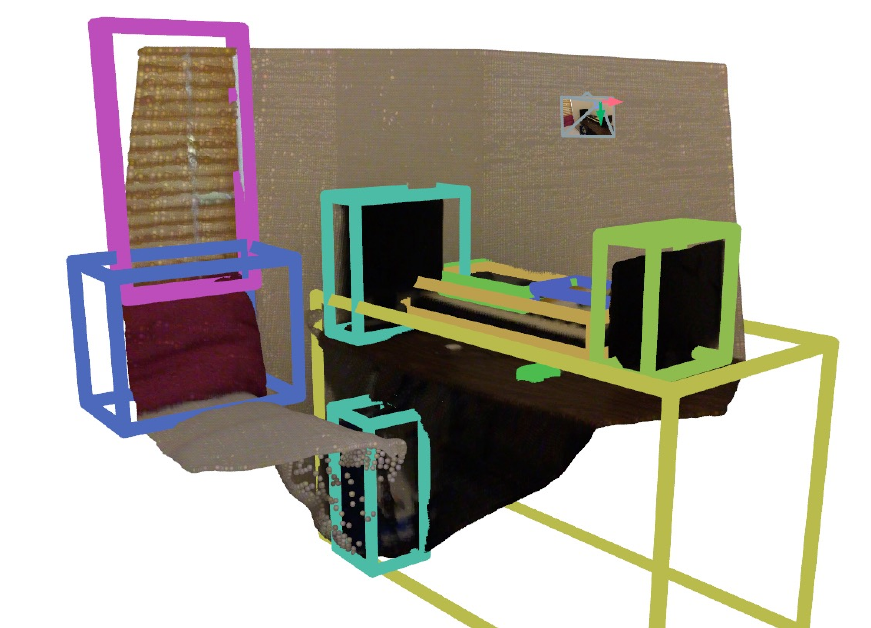} \\
    \includegraphics[width=\szi\linewidth]{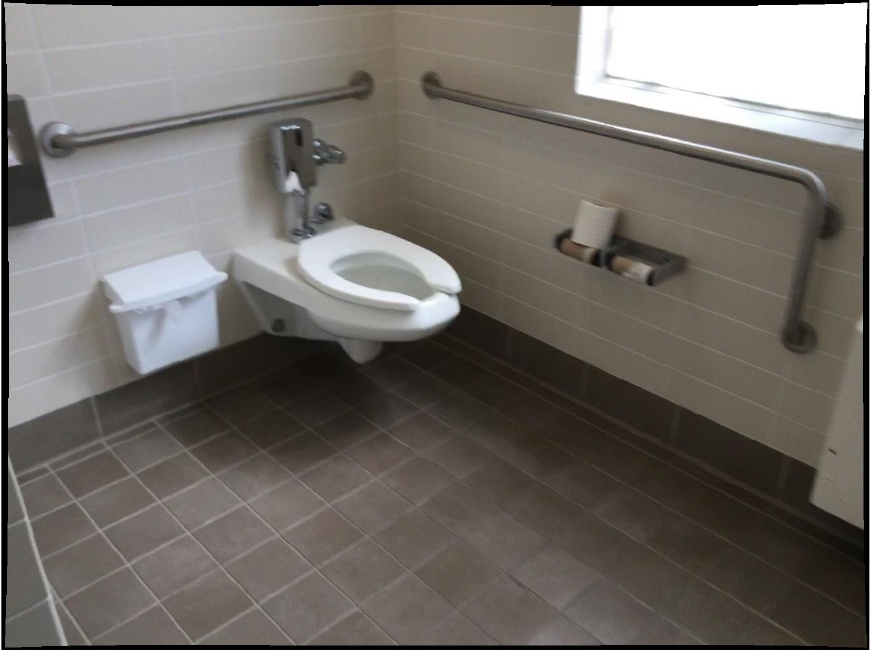} &
    \includegraphics[width=\szi\linewidth]{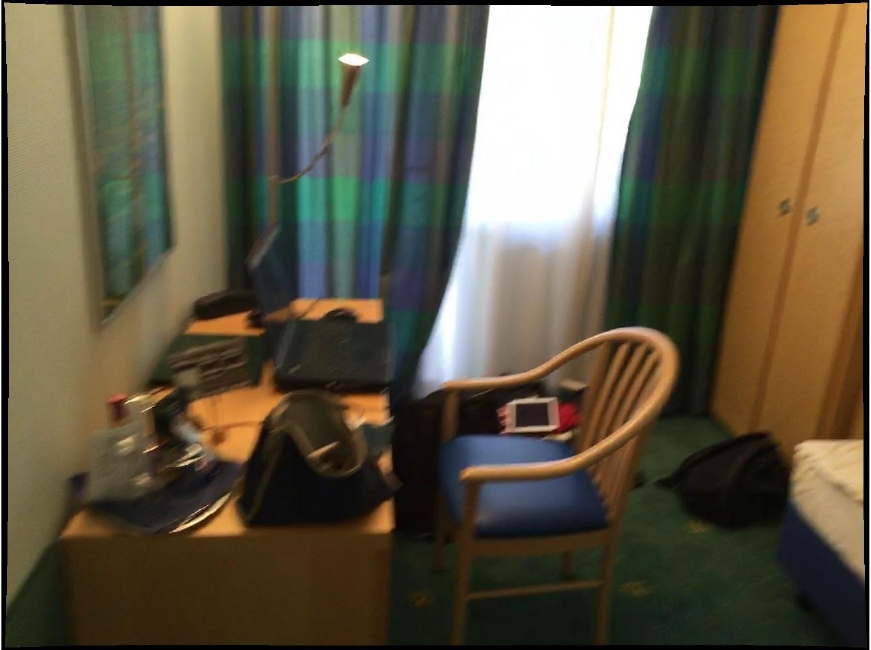} & \includegraphics[width=\szo\linewidth]{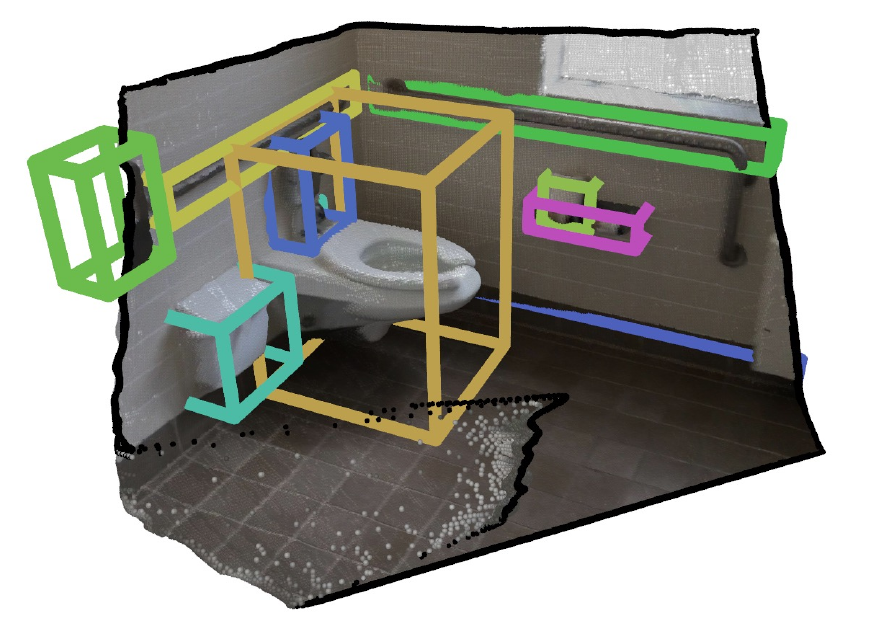} & \includegraphics[width=\szo\linewidth]{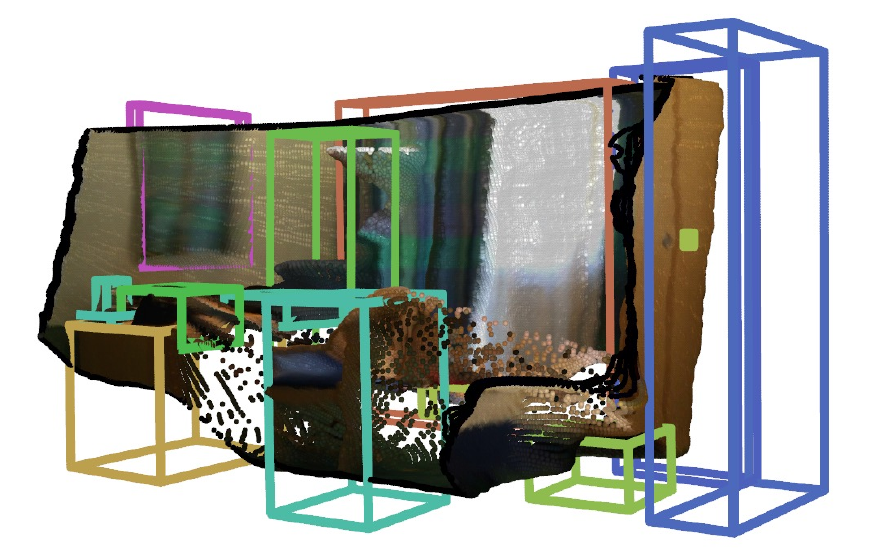} \\
    \multicolumn{2}{c}{RGB Input} & \multicolumn{2}{c}{3D Output} \\
    \end{tabular}
    \caption{
        \textbf{Qualitative Results on ScanNetV2.}
        We visualize the 3D bounding boxes plus depth estimation from \ourmodel under the zero-shot settings.
    }
    \label{fig:qualitative_scannet}
\end{figure}

%% file: sec/05_Conclusion.tex
\section{Limitations}
\label{sec:limit}
Despite the fact that we are using MapAnything as our detection transformer encoder, given the limitation of resources, we only train \ourmodel on CA-1M.
This leads to the limitation that \ourmodel works well only for indoor scenes.

Another constraint is the categorization of detected objects: since we focus on the geometric challenge of 3DOD from images, \ourmodel's detections are not put in relation to any closed or open-vocabulary queries. We leave open-vocabulary as future work to be explored as a natural extension of our proposed class-agnostic 3DOD model using the FF3R prior. 
Potentially, the class-agnostic detections from \ourmodel can also be semantically matched to queries using simple set-of-marks prompting in the input views with a VLM~\cite{yang2023set}.

\section{Conclusion}
\label{sec:conclusion}
In this paper, we introduce \ourmodel, a novel framework for online multi-view 3D object detection from streaming monocular inputs.
We demonstrate that adapting a metric feed-forward 3D reconstruction as a geometric prior effectively addresses the inherent depth and scale ambiguities that traditionally limit monocular vision.
By repurposing FF3R as the detection transformer encoder, we bypass the conventional 2D-to-3D lifting paradigm and propose an up-to-scale 3D detection head that optimally consumes these strong geometric priors to directly predict accurate metric-scale 3D bounding boxes and yield better performance.
\ourmodel achieves new SOTA performance on both in-domain and out-of-domain benchmarks, demonstrating our method's ability to successfully detect indoor objects using strong FF3R priors.

\section*{Acknowledgement}
This research was partially funded by the ETH AI Center Postdoc Fellowship (S. Hong), ETH Foundation Project 2025-FS-352 (Z. Bauer),  the SNSF Advanced Grant 216260 (Z. Bauer), Swiss AI Initiative from the Swiss National Supercomputing Centre (CSCS) grant under project ID a144 (Y-H. Yang), and the Lamarr Institute for Machine Learning and Artificial Intelligence (H. Blum).